\documentclass[sigconf]{acmart}

\usepackage{subfig}
\usepackage{multirow}
\usepackage{graphicx}
\usepackage{enumitem}
\usepackage{balance}
\usepackage{algorithm}
\usepackage{algorithmic}
\usepackage{hyperref}
\usepackage{cleveref}
\usepackage{comment}
\newcommand{\proposed}{\textsf{DefaultGNN}}
\newcommand\HH{
  \global\let\savedtextbullet\textbullet
  \gdef\textbullet{%
    \par\noindent\savedtextbullet\global\let\textbullet\savedtextbullet
  }%
}
\AtBeginDocument{%
  }

\copyrightyear{2026}
\acmYear{2026}
\setcopyright{cc}
\setcctype{by}
\acmConference[CIKM '26]{Proceedings of the 35th ACM International Conference on Information and Knowledge Management}{November 07--11, 2026}{Rome, Italy}
\acmBooktitle{Proceedings of the 35th ACM International Conference on Information and Knowledge Management (CIKM '26), November 07--11, 2026, Rome, Italy}
\acmDOI{10.1145/3799682.3840153}
\acmISBN{979-8-4007-2539-5/2026/11}

\begin{document}

\title[DefaultGNN]{DefaultGNN: A Dual-Perspective GNN Framework for Predicting Corporate Default from Buyer-Seller Transaction Networks}

\author{Junghoon Kim}
\affiliation{%
  \institution{KAIST}
  \city{Daejeon}
  \country{Republic of Korea}
  }
\email{jhkim611@kaist.ac.kr}

\author{Hyunsung Kim}
\affiliation{%
  \institution{KAIST}
  \city{Daejeon}
  \country{Republic of Korea}}
\email{hyunsung.kim@kaist.ac.kr}

\author{Seungyoon Choi}
\affiliation{%
  \institution{KAIST}
  \city{Daejeon}
  \country{Republic of Korea}}
\email{csyoon08@kaist.ac.kr}

\author{KyoungYong Park}
\affiliation{%
  \institution{Techfin Ratings}
  \city{Seoul}
  \country{Republic of Korea}}
\email{hess\_kpark@techfinratings.com}

\author{Jihun Lee}
\affiliation{%
  \institution{Techfin Ratings}
  \city{Seoul}
  \country{Republic of Korea}}
\email{jihuny@techfinratings.com}

\author{YongGu Ji}
\affiliation{%
  \institution{Douzone}
  \city{Seoul}
  \country{Republic of Korea}}
\email{todcode@douzone.com}

\author{Chanyoung Park}
\authornote{Corresponding author.}
\affiliation{%
  \institution{KAIST}
  \city{Daejeon}
  \country{Republic of Korea}}
\email{cy.park@kaist.ac.kr}

\renewcommand{\shortauthors}{Junghoon Kim et al.}

\begin{abstract}
Corporate default prediction is a core problem in financial risk management, yet traditional credit models rely heavily on financial statements that are often sparse or unavailable for many firms. Corporate transaction networks offer a complementary view of real economic activity, but how risk propagates through buyer–seller relationships remains underexplored. We conduct a large-scale empirical study using real-world electronic tax-invoice data spanning six years that links transaction histories with default events, revealing that transaction-driven risk is both role-dependent (buyer or seller) and scale-dependent. Based on these findings, we construct multiplex buyer-view and seller-view transaction networks and propose \proposed, a dual-perspective graph neural network-based framework for corporate default prediction. \proposed~integrates both views to model how risk flows through transactional relationships, achieving strong improvements over both attribute-based and graph-based baselines, especially for firms with limited intrinsic risk signals. We further provide interpretable network-based explanations by visualizing how distressed trading partners contribute to default risk. In collaboration with a licensed credit rating agency, we validate that \proposed's predictions complement existing credit scoring models, improving approval rates by 7--11\%p without increasing default risk among approved firms. The source code can be found at \url{https://github.com/jhkim611/DefaultGNN}.
\end{abstract}




\maketitle

\vspace{-3ex}
\section{Introduction}
\looseness=-1
\textit{Corporate credit evaluation} (or \textit{default prediction}), the task of determining whether a firm will default on its financial obligations in the future, plays a crucial role in financial decision-making, including lending decisions and inter-corporate trade activities. Traditional corporate credit scoring models primarily rely on financial statements and historical default data~\cite{financialratios1, financialratios2}, typically using statistical methods like logistic regression~\cite{altmanz, ohlsono, logreg, creditscoring, statmodel}. While these approaches are interpretable have shown stable performance, they are inherently constrained by the availability and timeliness of financial data~\cite{altdata1, altdata2}.

This limitation is particularly pronounced for small and medium enterprises (SMEs) and sole proprietors, who typically face infrequent financial reporting cycles and limited disclosure requirements. As a result, financial statements for these firms are often missing, outdated or incomplete, leading to information asymmetry between financial institutions and the firms being evaluated. In such settings, traditional financial statement-driven credit scoring models tend to overestimate risk for SMEs\footnote{Such "credit rationing" occurs when financial institutions impose limits on loan exposure due to difficulties in controlling risk through interest rate adjustments.}
~\cite{creditrationing1, creditrationing2, creditrationing3}, motivating growing interest in machine learning-based models that incorporate non-financial data in credit assessment systems.

Among such non-financial data, transactional relationships between businesses have garnered attention as they directly reflect the actual economic activity of firms. Businesses do not operate in isolation; they are interconnected through buyer-seller relationships. The default of one partner can cascade through a transaction network, impacting the financial stability of adjacent firms. Prior research has shown that credit risk contagion, i.e., the transfer of financial risk through transactions, can significantly influence default probability, with a meaningful correlation between a firm’s credit risk and that of its trading partners~\cite{contagion1, contagion2, contagion3, tradecredit1, tradecredit2}.

Despite this, large-scale studies that leverage corporate transaction data for default prediction remain limited, as traditional statistical or linear models cannot effectively incorporate the relational and multi-layered structure of interfirm transaction networks. Recent advances in graph neural networks (GNNs)~\cite{gcn, gat} have enabled relational modeling of financial systems, and have been applied to settings such as guarantee networks and supply chains for credit risk prediction~\cite{supplychaingnn, dgann, prev2}. However, existing GNN-based studies have largely underutilized transactional information. In particular, most models either focus on non-transactional relations (e.g., loans or guarantees) or simplify transactions and their accompanying risks to unidirectional and binary links, ignoring their bidirectional risk exposure and heterogeneous scale.

\looseness=-1
Specifically, in real-world transactions, risk would propagate in both directions: \textit{sellers} face exposure to late or failed payments, while \textit{buyers} are vulnerable to delivery failures that can disrupt downstream operations and other business relationships. Moreover, larger transactions would likely induce higher risk than smaller ones. Capturing these effects requires a more detailed analysis of how buyer–seller relationships and transaction magnitude shape default risk.

To address this gap, we conduct a large-scale empirical analysis of real-world electronic tax-invoice data, linking transaction histories with default events of trading partners (see Section~\ref{analysis}). We examine how default risk depends on who trades with whom, in which role (buyer or seller), and at what scale, and use these findings to construct multiplex buyer- and seller-view transaction networks. These insights are then incorporated into \textbf{\proposed}, a dual-perspective GNN framework for predicting corporate default from buyer-seller transaction networks. \proposed~ integrates both views, i.e., transactional patterns learned from separate GNN layers, to predict corporate default, enabling more accurate risk assessment especially for firms with limited intrinsic financial risk signals.

Overall, our contributions can be summarized as follows:
\vspace{-2ex}
\begin{itemize}[leftmargin = 2mm]
  \item \textbf{Large-scale empirical study of transaction-driven default risk.} We analyze real-world electronic tax-invoice data spanning six years that link buyer-seller transactions with default events, providing direct evidence for how risk propagates through corporate transaction networks.  
  \item \textbf{Role- and scale-aware modeling of transaction-based risk.} We show that default risk depends on buyer/seller roles and transaction magnitude, and construct multiplex transaction networks that explicitly capture these nuances.
  \item \textbf{Effectiveness.} Our proposed \proposed~integrates buyer-view and seller-view embeddings to model how risk is transmitted through transactional relationships, enabling more accurate default prediction, especially for firms with limited financial data.
  \item \textbf{Interpretable network-based explanations of default risk.} \proposed ~ enables visualization of a firm's transaction neighborhood, revealing how defaulted or distressed trading partners contribute to the predicted risk of the target firm, providing actionable insights for financial decision-making.
  \item \textbf{Practical validation with a credit rating agency.} In collaboration with Techfin Ratings, we validate that \proposed's predictions complement an existing credit scoring model, improving approval rates without increasing default risk.
\end{itemize}

\vspace{-2ex}
\section{Related Works}
\subsection{Modeling for Corporate Default Prediction}
Corporate default prediction has traditionally relied on financial statement-based models, where bankruptcy risk is estimated using accounting ratios and firm-level attributes. Classical approaches such as the Altman Z-score~\cite{altmanz} and the Ohlson O-score~\cite{ohlsono} remain widely used benchmarks due to their interpretability and strong theoretical grounding. However, their applicability depends critically on the availability of detailed financial statements, which are often missing, delayed or unavailable for small and medium enterprises (SMEs) and private firms~\cite{creditrationing1, creditrationing2, creditrationing3}.

\looseness=-1
To address these limitations, subsequent studies have explored machine learning-based credit models, including logistic regression and tree-based methods~\cite{logreg, xgboost, creditscoring}, which capture nonlinear relationships among firm attributes. While these often improve predictive performance, they typically treat firms as independent entities and do not explicitly model inter-firm relationships, overlooking the fact that default risk can propagate through economic connections.

\vspace{-2ex}
\subsection{Network-Based Financial Risk and GNNs}
Recognizing that firms operate within interconnected economic systems, prior research has studied default risk contagion through inter-firm networks, including trade relationships, supply chains and loan-guarantee structures. These studies provide empirical evidence that a firm’s default risk is correlated with that of its business partners, highlighting the importance of relational information~\cite{contagion1, contagion2, contagion3, supplychaingnn, supplychain}.

More recently, graph neural networks (GNNs)~\cite{gcn, gat, prev2} have enabled scalable learning on relational financial data and have been applied to settings such as loan-guarantee networks and supply-chain risk assessment, often outperforming traditional attribute-based models~\cite{financegraph1, financegraph2, financegraph3, financegraph4}. Notably, DGANN~\cite{dgann} leverages guarantee relationships to model default risk in lending systems.

\looseness=-1
Despite these advances, existing network analyses and GNN-based approaches often focus on non-transactional relations or represent transactions as single-view, unweighted and unidirectional graphs, overlooking key characteristics of real-world corporate transactions. In particular, buyer-seller perspectives and transaction magnitude, which are critical to understanding how risk propagates through transaction networks, remain largely underexplored. Our work addresses this gap by combining large-scale empirical analysis of transaction data with role- and scale-aware network modeling.

\vspace{-1ex}
\section{Data Description and Analysis} \label{analysis}
\subsection{Data Collection and Preprocessing}
The datasets used in this study consist of firm-level attributes, inter-firm transaction records, and corporate default information spanning six years (2018--2023). Firm attributes and transaction data were obtained from a large-scale electronic tax-invoice system operated by Douzone Bizon, which is widely used by businesses in Korea for statutory reporting. Corporate default labels were integrated based on credit event records compiled by a licensed credit rating agency in accordance with Basel II default definitions. Preprocessing steps include removing firms with invalid identifiers or missing business type, discarding non-positive transaction amounts, and aggregating multiple transactions between the same pair of firms within each month by transaction role, retaining at most two directed interactions per firm pair (each firm acting as the seller) with transaction amounts summed accordingly. Data collection and usage were conducted within a secure, access-controlled environment provided by the data owner.

\vspace{-1ex}
\subsection{Default-Transaction Correlation}
\subsubsection{Effect of Partner Default}
\begin{figure}[t]
\centering
\includegraphics[width=0.8\linewidth]{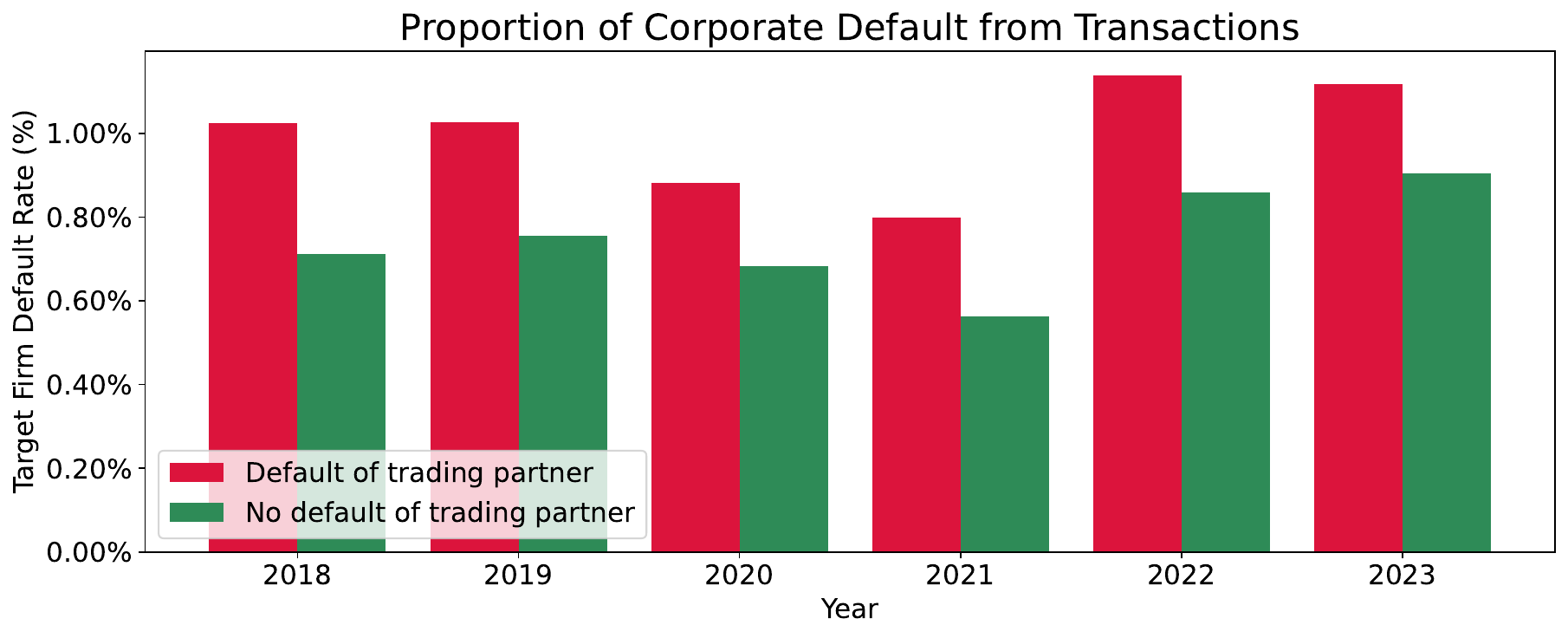}
\vspace{-3ex}
\caption{Transactions involving partners with default history are more likely to cause future default of the target firm.}
\vspace{-3ex}
\label{fig:default1}
\end{figure}

We first examine how the default history of trading partners relates to a firm’s future default risk. At the transaction level, transactions involving partners with prior (i.e., in the previous 12 months) default history are consistently associated with a higher likelihood of future (i.e., in the next 12 months) default of the target firm across all years (see Fig.~\ref{fig:default1}).

\begin{figure}[t]
\centering
\includegraphics[width=0.8\linewidth]{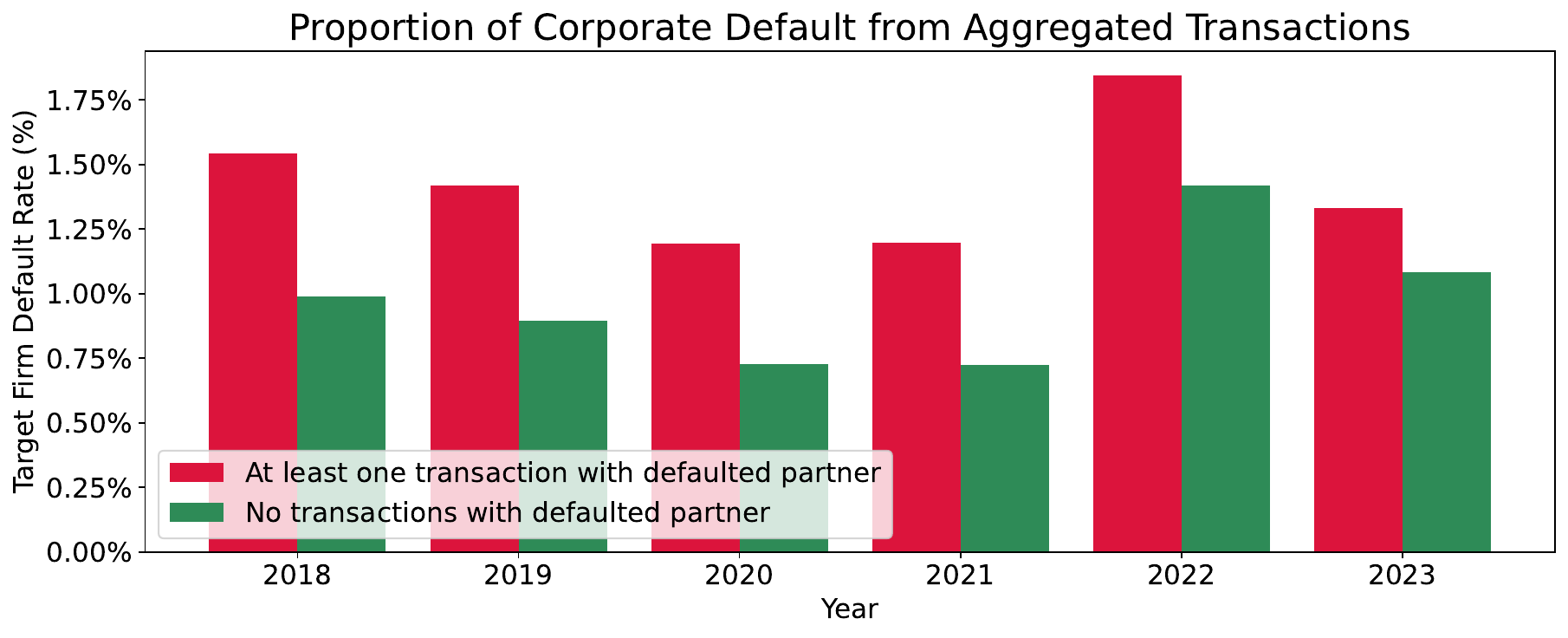}
\vspace{-3ex}
\caption{Firms with defaulted trading partners exhibit higher future default rates than those whose partners have no prior default history.}
\vspace{-3ex}
\label{fig:default2}
\end{figure}

\begin{figure}[t]
\centering
\includegraphics[width=0.95\linewidth]{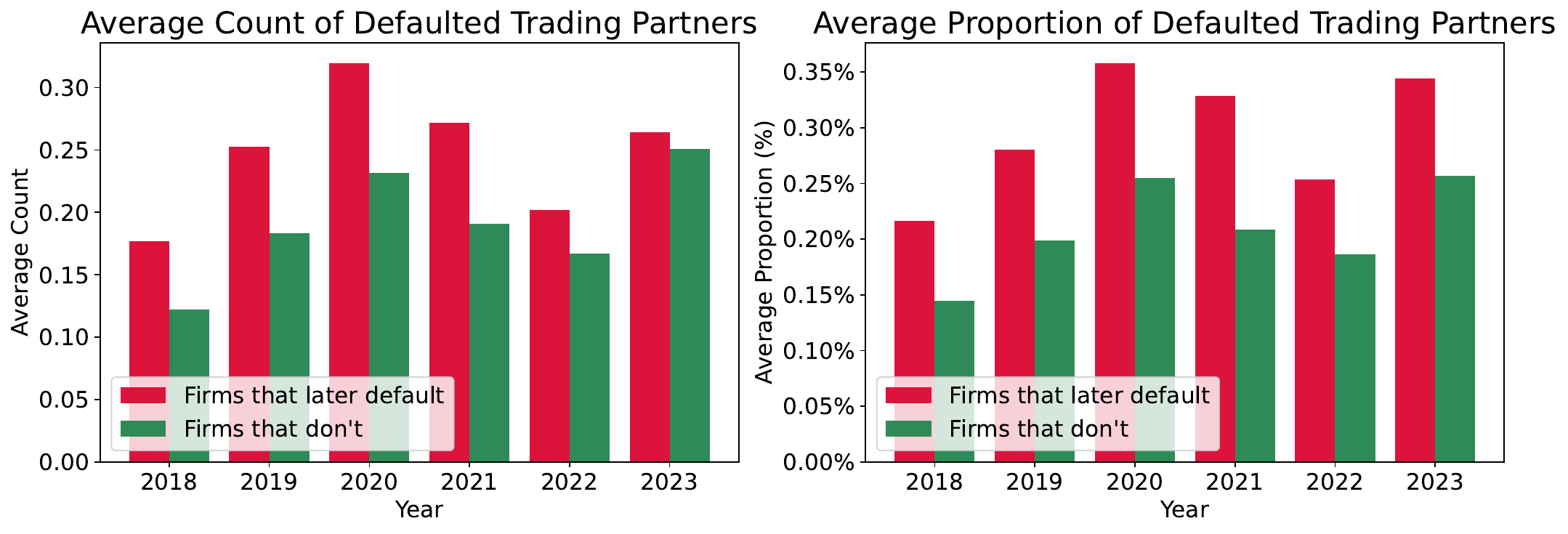}
\vspace{-3ex}
\caption{Firms that later default tend to have more defaulted trading partners.}
\vspace{-4ex}
\label{fig:default3}
\end{figure}

This pattern persists when aggregating transactions at the firm level. As shown in Fig.~\ref{fig:default2}, firms that transact with at least one defaulted partner exhibit substantially higher future default rates than those whose partners have no default history. Moreover, firms that eventually default tend to have both a higher number and a higher proportion of defaulted trading partners among their relationships (see Fig.~\ref{fig:default3}). These trends remain consistent across time.

Overall, these results indicate a \textbf{strong correlation between partner default history and future corporate default}, underscoring the importance of modeling inter-firm dependencies.

\subsubsection{Effect of Transaction Role}
A transaction can be viewed from two complementary perspectives: as a sale for the seller and as a purchase for the buyer. While prior work often models default propagation in a single direction — typically emphasizing payment risk faced by sellers, analogous to loaners exposed to guarantee risk~\cite{dgann} — such a unidirectional view overlooks risk exposure on the buyer side, where delivery failures or operational disruptions can also have cascading effects.

\begin{figure}[t]
\centering
\includegraphics[width=0.8\linewidth]{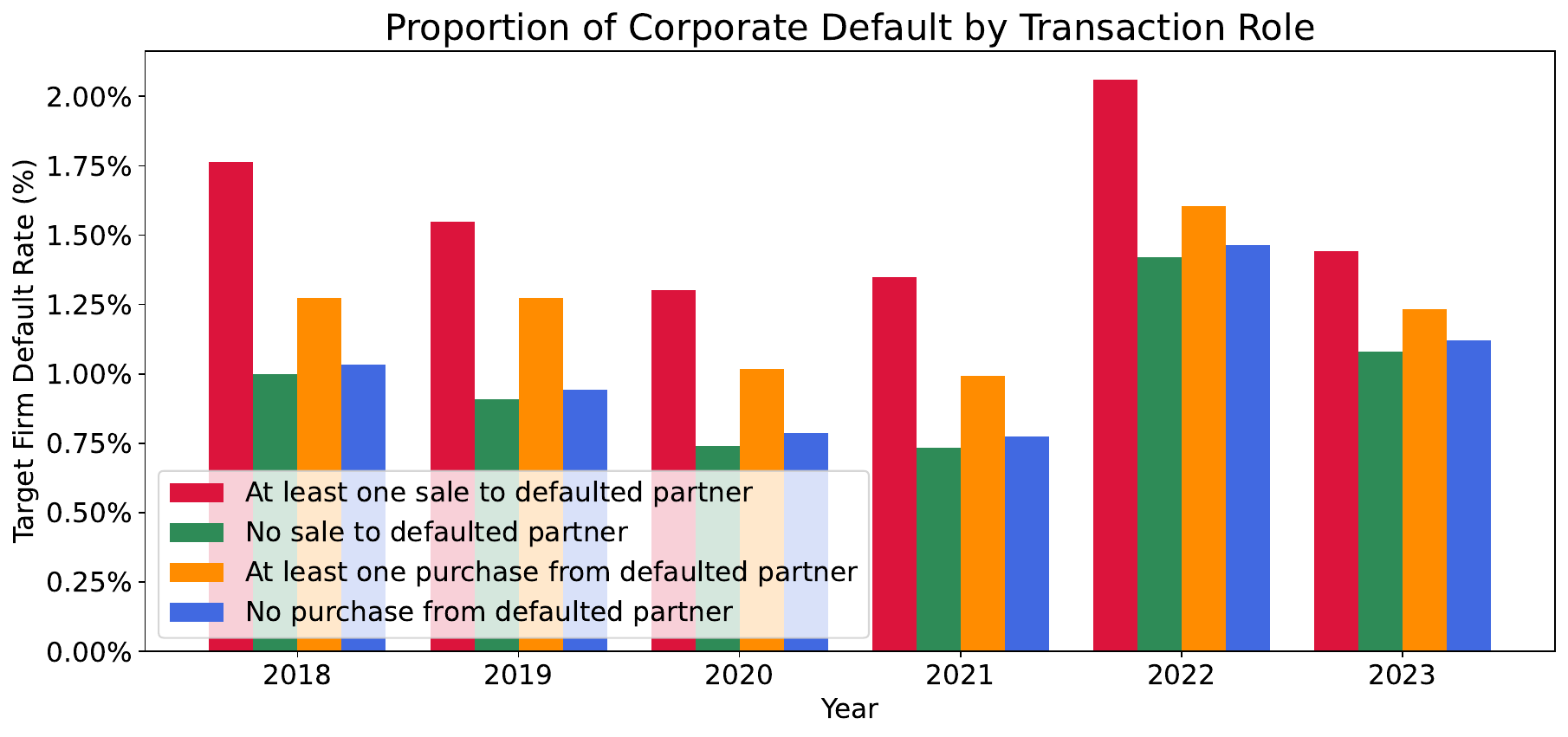}
\vspace{-3ex}
\caption{\textit{Both} sales to and purchases from defaulted partners are associated with increased future default risk, indicating bidirectional risk propagation.}
\vspace{-3ex}
\label{fig:default5}
\end{figure}

Empirical results support the relevance of both perspectives. As shown in Fig.~\ref{fig:default5}, both sales to and purchases from defaulted partners are associated with elevated future default risk. This suggests that \textbf{default risk propagates through transactions in a bidirectional manner}, motivating the need to explicitly distinguish buyer-view and seller-view transaction networks.

\begin{figure}[t]
\centering
\includegraphics[width=0.8\linewidth]{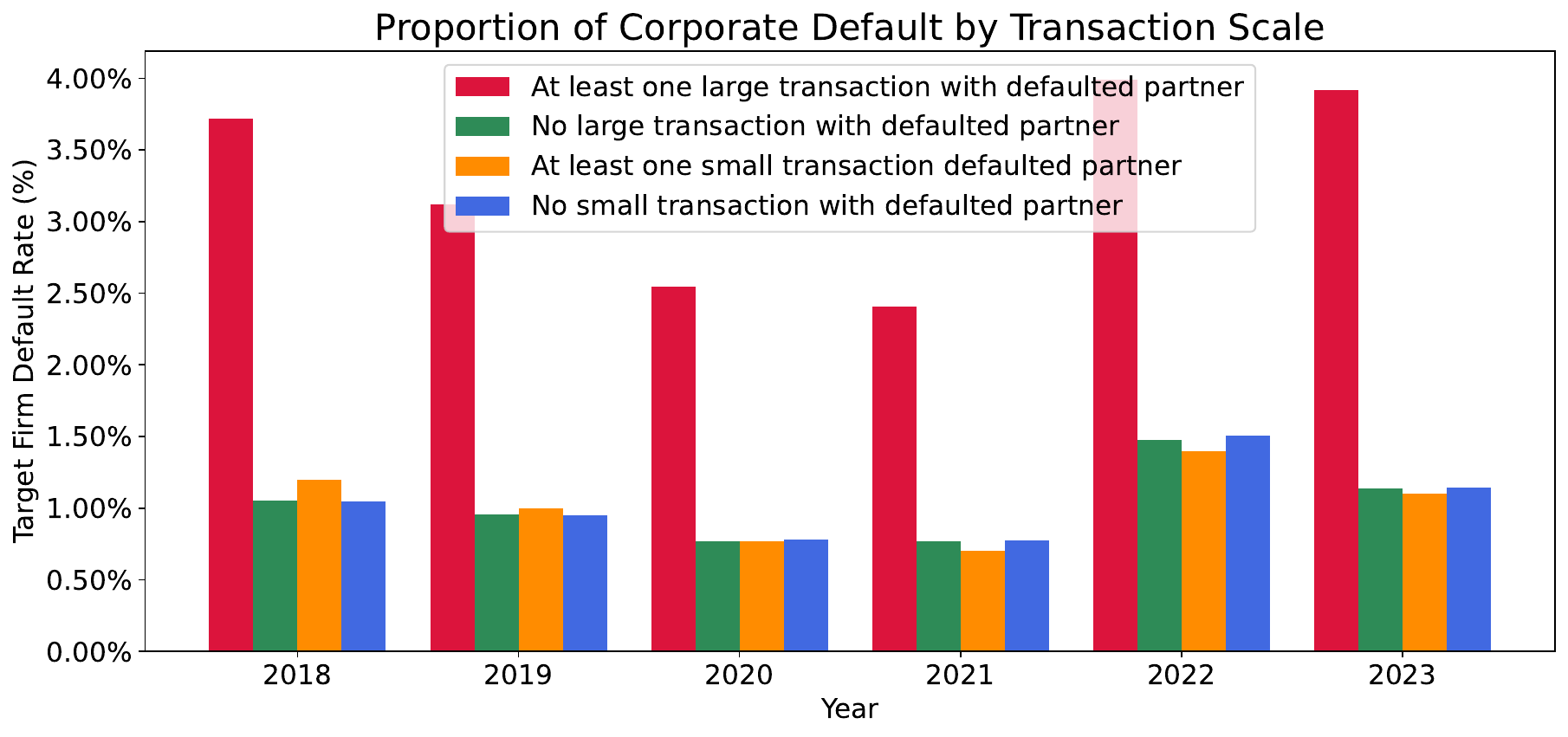}
\vspace{-3ex}
\caption{Large-scale (proportion $\ge 0.5$) transactions with defaulted partners contribute substantially more to future default risk than small-scale (proportion $< 0.5$) transactions.}
\vspace{-3ex}
\label{fig:default6}
\end{figure}

\subsubsection{Effect of Transaction Scale}
We further investigate how transaction magnitude influences default propagation. Transaction scale is defined as the ratio of a transaction’s amount to the firm’s total transaction volume within the same month.

As shown in Fig.~\ref{fig:default6}, firms that engage in large-scale transactions with defaulted partners are significantly more likely to experience future default than firms without such transactions. In contrast, small-scale transactions with defaulted partners exhibit a much weaker effect. This demonstrates that \textbf{transaction magnitude plays a critical role in shaping default risk} and should be explicitly incorporated into relational modeling.

\vspace{-1ex}
\subsubsection{Summary of Analysis}
In summary, our empirical analysis identifies three key factors underlying default propagation in transaction networks: 1) the default history of trading partners, 2) the transactional role of firms as buyers or sellers, and 3) the relative scale of transactions. These results indicate that default risk propagates through transactions in a bidirectional and heterogeneous manner, with economically significant transactions contributing disproportionately to future default risk. All observed trends remain consistent across the years considered.

These findings highlight the limitations of unidirectional or binary relational modeling and motivate a dual-perspective, edge-weighted graph-based approach that explicitly accounts for heterogeneous risk exposure.

\vspace{-2ex}
\subsection{Multiplex Transaction Networks} \label{construction}
Based on the transactional data, we construct multiplex inter-firm transaction networks that explicitly distinguish buyer–seller roles and transaction scale. For each year, firms are represented as nodes, and transactions between firms are represented as directed edges. To reflect asymmetric risk propagation mechanisms, we define two complementary transaction views over the same set of firms.

In the \textbf{seller-view} transaction network, a directed edge from firm $u$ to firm $v$ represents a transaction where $v$ acts as the seller and $u$ acts as the buyer. This view captures risk exposure arising from delayed or failed payments, where financial stress at the buyer can propagate upstream to the seller. Conversely, in the \textbf{buyer-view} transaction network, a directed edge from $v$ to $u$ represents the same transaction, reflecting the buyer’s exposure to delivery failures or operational disruptions at the seller.

To account for heterogeneous economic impact, each edge is assigned a weight reflecting the relative transaction scale. Specifically, for a transaction between firms $u$ and $v$, the edge weight is defined as the ratio of the transaction amount to the total sales (seller view) or total purchases (buyer view) of the corresponding firm within the same period, i.e., month. To mitigate the effect of extreme values, the weights can be further scaled by the following function: $w_{uv}^{new} = log(1+\alpha\cdot w_{uv}^{orig})/log(1+\alpha)$, where $w_{uv}^{orig}$ denotes the initial relative transaction ratio and $\alpha$ is a scaling factor.

The resulting multiplex representation consists of two directed, edge-weighted graphs sharing the same node set but encoding distinct transactional semantics (see Table~\ref{table:stats} for summary statistics). By jointly modeling these buyer- and seller-view transaction networks, this representation captures role- and scale-dependent risk propagation patterns inherent in real-world corporate transactions.

\vspace{-2ex}
\subsection{Firm Attributes}
\looseness=-1
Each firm is represented by a set of non-financial firm attributes (108-dimensional) that capture basic operational characteristics under realistic data constraints where detailed financial statements are unavailable. These include a recent (within that year) default indicator, taxation category, business type encodings at multiple levels of granularity, and log-scaled aggregate sales and purchase statistics computed from historical transaction records. Importantly, these features do not encode information about specific trading partners or network structure, serving as a baseline that is substantially enhanced by the relational information from transaction networks.

\vspace{-3ex}
\section{Proposed Framework:~\proposed} \label{method}
\subsection{Problem Setting and Overview}
\looseness=-1
Let $\mathcal{D = (V, E)}$ be a yearly corporate transaction dataset. $\mathcal{V}$ is a set of nodes, each node corresponding to an individual firm. $x_v \in \mathbb{R}^F$ denotes the node features of each node $v \in \mathcal{V}$, consisting of $F$ firm attributes. Each node is further labeled as $y_v \in \{0,1\}$ based on whether the corresponding firm defaults within the following year, i.e., $y_v = 1$ for defaulted firms/nodes and $y_v = 0$ for non-defaulted firms. $\mathcal{E} = [E^{sell}, E^{buy}]$ contains two edge sets, corresponding to the seller-view and buyer-view transaction networks defined in Section~\ref{construction}, respectively. Specifically, each directed edge $(u, v)$ in $E^{view}$ ($view \in \{sell, buy\}$) is paired with edge weight $w_{uv}$, where the directions and weights in each set are defined to represent their respective view. In other words, $\mathcal{D}$ contains year-wise multiplex transaction networks for a set of firms. For clarity, we additionally denote $\mathcal{G}^{sell}=(\mathcal{V}, E^{sell})$ and $\mathcal{G}^{buy}=(\mathcal{V}, E^{buy})$ as the seller- and buyer-view graphs, respectively, where $\mathcal{V}$ is shared among the two graphs.

Given dataset $\mathcal{D = (V, E)}$, our objective is to learn a binary classifier for corporate default prediction. In detail, the classifier is first trained on a training subset $\mathcal{V}_{train}$ and validated on a validation subset $\mathcal{V}_{val}$, finally evaluated by predicting the node labels in a test subset $\mathcal{V}_{test}$, whose labels were not available during training.

\begin{figure}[t]
\centering
\includegraphics[width=0.85\linewidth]{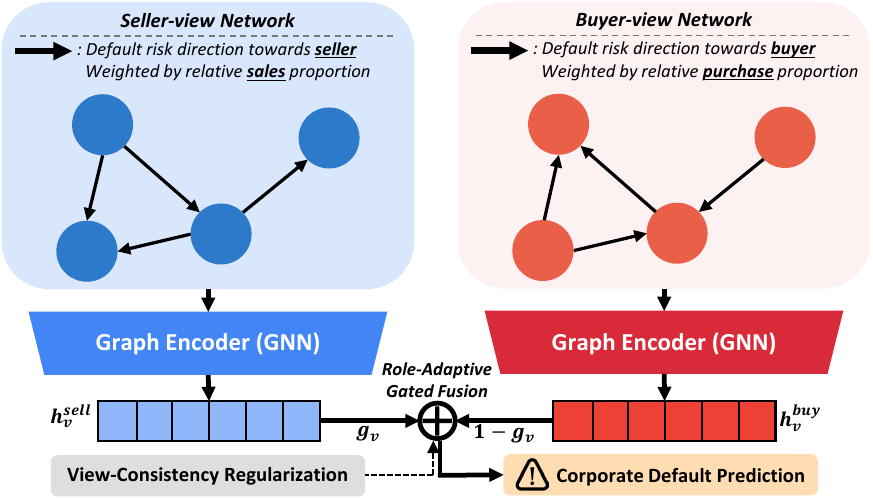}
\vspace{-3ex}
\caption{Overall framework of ~\proposed. View-specific embeddings learned from multiplex transaction networks are fused via gating, regularized through view consistency. The fused embeddings are then used for default prediction.}
\vspace{-4ex}
\label{fig:architecture}
\end{figure}

Our proposed \textbf{~\proposed} is a dual-perspective GNN-based framework (see Fig.~\ref{fig:architecture}) that \textbf{(1)} learns view-specific representations from each transaction network, \textbf{(2)} integrates them via a role-adaptive gating mechanism further stabilized through a view-consistency regularizer, and finally \textbf{(3)} utilizes the fused embeddings to identify future defaults of nodes.

\vspace{-2ex}
\subsection{View-Specific Graph Encoders}
\proposed~ employs two parallel graph encoders, one for each transaction view. Both encoders share the same architectural design but operate on different graph structures, and map firms and their relationships into latent representations that capture default risk propagation patterns specific to the corresponding transaction role.

Specifically, each encoder consists of multiple GNN layers that perform structured message passing over the transaction graph, aggregating information from trading partners while accounting for transaction scale represented through edge weights. Formally, node representation for a firm node $v$ at layer $l$ is updated as:
\vspace{-2ex}
\begin{equation} \label{eqn:gnn}
    h_v^{(l+1)} = \phi \left( \sum_{u \in \mathcal{N}(v) \cup (v)} \alpha_{uv}(w_{uv})\mathbf{W}^{(l)}h_u^{(l)} \right),
\end{equation}
where $h_u^{(l)}$ denotes the representation of node $u$ ($h_u^{(0)} = x_u$ initially), $\mathcal{N}(v)$ is the set of 1-hop neighbor nodes of node $v$, $\mathbf{W}^{(l)} \in \mathbb{R}^{d \times d}$ is the learnable weight matrix ($d$ is the latent embedding dimension size and $\mathbf{W}^{(0)} \in \mathbb{R}^{d \times F}$), $w_{uv}$ is the edge weight corresponding to directed edge $(u,v)$, $\alpha_{uv}(w_{uv})$ is a normalized aggregation coefficient that depends on both graph structure and transaction scale, and $\phi(\cdot)$ is a non-linear activation. By incorporating transaction magnitude directly into message passing, the encoders capture heterogeneous risk exposure arising from economically significant trading relationships, consistent with our empirical analysis. After the final GNN layer, the resulting representations are taken as the view-specific embeddings: $h_v^{sell}$ and $h_v^{buy}$.

\subsection{Role-Adaptive Gated Fusion}
Given the seller- and buyer-view embeddings $h_v^{sell}$ and $h_v^{buy}$, \proposed~ integrates them using a gating mechanism that adaptively determines their relative importance on a per-firm basis. The fusion gate is computed as ($\sigma(\cdot)$ is the sigmoid function):
\begin{equation} \label{eqn:gate1}
g_v = \sigma(w_{gate_v}^\intercal[h_v^{sell}||h_v^{buy}] + b).
\end{equation}
The gated embeddings are computed as $\tilde{h}_v^{sell} = g_v \cdot h_v^{sell}$ and $\tilde{h}_v^{buy} = (1-g_v) \cdot h_v^{buy}$, respectively, and concatenated to form the fused representation $z_v = [\tilde{h}_v^{sell}||\tilde{h}_v^{buy}]$. $z_v$ is then passed to a linear classifier to predict default probability. This gating mechanism allows ~\proposed~ to account for heterogeneity in transactional roles, enabling the model to emphasize buyer-side or seller-side risk exposure depending on the firm’s position in the transaction network.

\smallskip
\noindent{\textbf{View-Consistency Regularization. }}
To promote stable integration of complementary transaction views, we introduce a \textit{view-consistency regularizer} that encourages compatible risk estimates when both views are confident, guiding the fusion process toward coherent representations.

Specifically, \proposed~additionally produces view-specific default probabilities: $p_v^{sell} = P(y=1|h_v^{sell}), p_v^{buy} = P(y=1|h_v^{buy})$, from which we can define view-specific confidences in range $[0,1]$: $c_v^{sell} = 2|p_v^{sell}-0.5|, c_v^{buy} = 2|p_v^{buy}-0.5|$.
Disagreement between the obtained probabilities is penalized through a confidence-weighted mean squared error:
\begin{equation} \label{eqn:reg3}
    \mathcal{L}_{cons} = \frac{\sum_v w_v^{c}(p_v^{sell}-p_v^{buy})^2}{\sum_v w_v^{c}}, w_v^{c} = c_v^{sell} c_v^{buy}.
\end{equation}

This regularization encourages agreement primarily when both views are confident, while allowing divergence when either view is uncertain. As a result, it stabilizes gated fusion without suppressing view-specific representations or forcing collapse.

\subsection{Training Objective}
\looseness=-1
\proposed~ is trained using weighted binary cross entropy loss to address the severe class imbalance inherent in our task, with far less firms experiencing future default (see Table~\ref{table:stats}). Formally, given prediction outputs $\hat{p}_v = \sigma(z_v)$ the classification loss is defined as:
\begin{equation} \label{eqn:pred}
    \mathcal{L}_{class} = -\sum_{v \in \mathcal{V}_{train}} (w_{bce} \cdot y_v \cdot log(\hat{p}_v) + (1-w_{bce}) \cdot (1-y_v) \cdot log(1-\hat{p}_v),
\end{equation}
where $w_{bce} \in [0,1]$ is assigned to give larger emphasis to correctly identifying defaulted firms.

The overall training objective is $\mathcal{L} = \mathcal{L}_{class} + \lambda \mathcal{L}_{cons}$, where $\lambda$ is a parameter controlling the strength of regularization. Through this procedure, \proposed~ effectively integrates both buyer- and seller-view transaction networks to predict future corporate default.

\section{Experiments}
In this section, we conduct comprehensive experiments to answer the following research questions:
\begin{itemize}[leftmargin = 2mm]
  \item \textbf{RQ1. } How well does our proposed \proposed~ perform in predicting corporate default compared with baselines?
  \item \textbf{RQ2. } How effective are our multiplex networks and additional modules in enhancing \proposed's performance?
  \item \textbf{RQ3. } Is \proposed~ robust under out-of-time settings?
  \item \textbf{RQ4. } Can \proposed~ identify and visualize distressed trading partners that critically contribute to a target firm's default?
\end{itemize}

\begin{table}[t]
\centering
\caption{Dataset statistics.}
\vspace{-3ex}
\resizebox{0.9\columnwidth}{!}{
\begin{tabular}{l c c c}
\hline
\textbf{Year} 
& \textbf{\# Nodes (Firms)} 
& \textbf{\# Defaulted Nodes (proportion)} 
& \textbf{\# Edges (Transactions)} \\
\hline
2018 & 811,866 & 8,480 (1.04\%) & 42,088,093 \\
2019 & 911,485 & 8,794 (0.96\%) & 52,693,636 \\
2020 & 950,377 & 7,646 (0.80\%) & 53,444,395 \\
2021 & 985,320 & 7,788 (0.79\%) & 54,823,028 \\
2022 & 930,853 & 13,718 (1.47\%) & 49,250,355 \\
2023 & 420,687 & 4,763 (1.13\%) & 14,175,876 \\
\hline
\end{tabular}
}
\label{table:stats}
\vspace{-4ex}
\end{table}

\vspace{-2ex}
\subsection{Experimental Setup}
\noindent{\textbf{Datasets. }}
We evaluate \proposed~on our constructed multiplex transaction networks (see Section~\ref{construction} for details) spanning six years, i.e., six distinct datasets. The statistics can be found in Table~\ref{table:stats}. Note that the statistics are equivalent for both buyer- and seller-view networks. 70\%, 10\% and 20\% of the firms are assigned to the training, validation and test sets, respectively, where the ratio of defaulted firms is kept consistent among them, e.g., for year 2021 they each  contain 5,451, 778 and 1,559 defaulted nodes (all 0.79\% of their corresponding sets), respectively.

\smallskip
\looseness=-1
\noindent{\textbf{Baselines.}} We compare against baselines spanning four categories. \textbf{Attribute-based methods (G1)} (Logistic Regression~\cite{logreg} and XGBoost~\cite{xgboost}) treat firms as independent instances, only utilizing firm attributes. We also include variants that incorporate additional transaction-related information as features ("+ Trans. Rel.": transaction counts with partners grouped by business type, and "+ Def. Agg.": monthly and annual counts of defaulted partners). \textbf{Standard GNNs (G2)} (GCN~\cite{gcn}, GAT~\cite{gat} and DGANN~\cite{dgann}: a directed graph attention network predicting defaults on loan-guarantee networks) model firms as nodes and transactions as edges, and are applied on the seller-view graph. \textbf{Multi-relational GNNs (G3)} (R-GCN~\cite{rgcn} and CompGCN~\cite{compgcn}) handle multiple edge types within a single model, operating on a combined graph where buyer and seller transactions are treated as distinct relation types. \textbf{Multiplex graph models (G4)} (DMGI~\cite{dmgi}, DMG~\cite{dmg} and MGHC~\cite{mghc}, adapted to the supervised setting) maintain separate encoders for each view-specific graph, mirroring \proposed's dual-encoder design.

\smallskip
\noindent{\textbf{Implementation Details.}} \proposed~is implemented in PyTorch 2.2.1 with CUDA 12.1 and trained on an NVIDIA RTX A6000 GPU. The view-specific graph encoders use a 2-layer GCN~\cite{gcn} with embedding dimension $d=128$. The edge scaling factor $\alpha$, consistency regularization weight $\lambda$, and BCE weight $w_{bce}$ are set to 50, 0.05, and 0.9, respectively. All parameters are optimized with Adam~\cite{adam} (learning rate 0.001) for up to 500 epochs with early stopping (patience 10). For all baselines, we follow the original implementations and suggested hyperparameters.

\begin{table*}[t]
\centering
\caption{Overall model performance. The best AR score for each dataset is highlighted in bold. Standard deviations for \proposed: $\pm$0.001--0.003 (All), $\pm$0.001--0.005 (NoHist), with comparable ranges for other methods. \proposed~significantly outperforms the best baseline on all datasets (Wilcoxon signed-rank test, $p = 0.016$).}
\vspace{-3ex}
\resizebox{0.8\textwidth}{!}{
\begin{tabular}{c l c c c c c c c c c c c c c c c c c c}
\hline
&\multirow{2}{*}{\textbf{Method}}&&\multicolumn{2}{c}{\textbf{2018}}&&\multicolumn{2}{c}{\textbf{2019}}&&\multicolumn{2}{c}{\textbf{2020}}&&\multicolumn{2}{c}{\textbf{2021}}&&\multicolumn{2}{c}{\textbf{2022}}&&\multicolumn{2}{c}{\textbf{2023}}\\
\cline{4-5}
\cline{7-8}
\cline{10-11}
\cline{13-14}
\cline{16-17}
\cline{19-20}
&&&All&NoHist&&All&NoHist&&All&NoHist&&All&NoHist&&All&NoHist&&All&NoHist\\
\hline
\multirow{6}{*}{\textbf{G1}}&Log. Reg.~\cite{logreg}&&0.378&0.324&&0.426&0.329&&0.431&0.315&&0.409&0.290&&0.358&0.257&&0.393&0.232\\
&\quad + Trans. Rel.&&0.369&0.313&&0.406&0.313&&0.410&0.297&&0.414&0.297&&0.362&0.260&&0.394&0.244\\
&\quad + Def. Agg.&&0.381&0.326&&0.420&0.324&&0.429&0.320&&0.412&0.295&&0.363&0.258&&0.396&0.246\\
&XGBoost~\cite{xgboost}&&0.382&0.294&&0.418&0.283&&0.419&0.253&&0.379&0.203&&0.428&0.294&&0.447&0.222\\
&\quad + Trans. Rel.&&0.378&0.289&&0.419&0.284&&0.424&0.260&&0.373&0.220&&0.434&0.302&&0.448&0.223\\
&\quad + Def. Agg.&&0.399&0.414&&0.419&0.284&&0.423&0.258&&0.389&0.210&&0.430&0.297&&0.456&0.234\\
\hline
\multirow{3}{*}{\textbf{G2}}&GCN~\cite{gcn}&&0.502&0.434&&0.531&0.430&&0.558&0.436&&0.540&0.417&&0.522&0.412&&0.575&0.410\\
&GAT~\cite{gat}&&0.499&0.431&&0.534&0.433&&0.552&0.439&&0.542&0.415&&0.518&0.408&&0.580&0.417\\
&DGANN~\cite{dgann}&&0.494&0.431&&0.513&0.411&&0.536&0.413&&0.533&0.402&&0.502&0.399&&0.565&0.404\\
\hline
\multirow{2}{*}{\textbf{G3}}&R-GCN~\cite{rgcn}&&0.509&0.432&&0.521&0.423&&0.542&0.419&&0.524&0.400&&0.515&0.404&&0.560&0.392\\
&CompGCN~\cite{compgcn}&&0.508&0.425&&0.528&0.434&&0.554&0.429&&0.523&0.398&&0.519&0.413&&0.561&0.395\\
\hline
\multirow{3}{*}{\textbf{G4}}&DMGI~\cite{dmgi}&&0.521&0.454&&0.549&0.457&&0.568&0.451&&0.550&0.436&&0.530&0.446&&0.589&0.430\\
&DMG~\cite{dmg}&&0.515&0.440&&0.536&0.445&&0.563&0.429&&0.551&0.440&&0.523&0.442&&0.583&0.421\\
&MGHC~\cite{mghc}&&0.518&0.457&&0.550&0.448&&0.557&0.443&&0.546&0.416&&0.538&0.435&&0.579&0.432\\
\hline
\textbf{Ours}&\proposed&&\textbf{0.542}&\textbf{0.482}&&\textbf{0.574}&\textbf{0.480}&&\textbf{0.586}&\textbf{0.473}&&\textbf{0.583}&\textbf{0.472}&&\textbf{0.558}&\textbf{0.463}&&\textbf{0.605}&\textbf{0.455}\\
\hline
\end{tabular}
}
\label{table:main}
\vspace{-3ex}
\end{table*}

\smallskip
\noindent{\textbf{Evaluation Details. }}
We evaluate model performance using the Accuracy Ratio (AR) metric derived from the Area Under the Curve (AUC) score, formally defined as $AR = 2 \times AUC - 1$ (range: [0, 1]). AR is a standard performance metric in credit risk modeling, widely used in industry~\cite{ar1, ar2, creditscoring} to assess the discriminatory power of default prediction models, especially under severe class imbalance where defaults are rare.

In practical credit assessment, many firms — especially SMEs and private firms — have no prior default history, limiting the effectiveness of models that rely on historical risk signals. To evaluate performance in such data-poor settings, we report the AR score not only on all test firms (\textbf{All}), but also on the subset of firms with no prior default history (\textbf{NoHist}). Performance on the latter reflects a model’s ability to infer emerging risk from relational transaction patterns rather than intrinsic historical indicators. For all experiments, we report the average performance of 3 independent runs.

\vspace{-2ex}
\subsection{Performance Comparison (RQ1)} \label{performance}
Our main results can be found in Table~\ref{table:main}. Attribute-based methods (G1) consistently underperform graph-based approaches by a large margin, even when augmented with transaction-related features, highlighting that encoding transactional data as simple features cannot capture the complex relational structure between firms.

\looseness=-1
Among graph-based methods, standard GNNs (G2) on the seller-view graph achieve strong results, demonstrating the value of relational modeling. However, multi-relational GNNs (G3) perform comparably to or even below standard GNNs, suggesting that merging buyer–seller signals within each GNN layer loses the view-specific information needed to capture asymmetric risk propagation.

\looseness=-1
Multiplex graph models (G4) consistently outperform both G2 and G3, confirming the importance of preserving view-specific representations. Nevertheless, \proposed~outperforms all G4 baselines across every year and both evaluation settings, despite sharing the same dual-encoder architecture. We attribute this to the domain-aligned gated fusion: \proposed's learned gate values exhibit a strong positive correlation (Pearson $r=0.73, p < 10^{-6}$) with firms' transaction role balance, measured by annual sales / (annual sales $+$ purchases). This indicates that the model adaptively emphasizes the seller- or buyer-view embedding depending on each firm's economic role, assigning higher weights to the seller view for seller-dominant firms and vice versa. In contrast, more generic fusion strategies such as averaging (DMGI) or disentanglement (DMG) do not capture this role-dependent structure.

\proposed's advantage is particularly pronounced for firms without prior default history (NoHist), the most practically important setting where traditional credit models fail. This reinforces that DefaultGNN effectively captures emerging risk from transactional relationships even in the absence of intrinsic historical risk signals.

\begin{table}[t]
\centering
\caption{Ablation studies on~\proposed~(AR score).}
\vspace{-3ex}
\resizebox{0.8\linewidth}{!}{
\begin{tabular}{c l|c c}
\hline
&&All&NoHist\\
\hline
\textbf{Ours}&\proposed&\textbf{0.583}&\textbf{0.472}\\
\hline
\multirow{4}{*}{\textbf{Network Variants}}&\proposed$+$SellerView&0.544&0.420\\
&\proposed+BuyerView&0.545&0.420\\
&\proposed+CollapsedView1&0.531&0.406\\
&\proposed+CollapsedView2&0.516&0.391\\
&\proposed$-$EdgeWeights&0.535&0.408\\
\hline
\multirow{3}{*}{\textbf{Fusion Variants}}&\proposed$-$Gating&0.565&0.444\\
&\proposed$-$ConsReg&0.560&0.437\\
&\proposed+CrossAttn&0.537&0.427\\
\hline
\end{tabular}
}
\label{table:ablation}
\vspace{-5ex}
\end{table}

\vspace{-3ex}
\subsection{Ablation Study (RQ2)} \label{ablation}
To assess the contribution of each component in \proposed, we conduct ablation experiments on the 2021 dataset (see Table~\ref{table:ablation}), with consistent trends observed across other years. \textbf{Network variants} isolate individual graph-related design choices: both \proposed+SellerView and \proposed+BuyerView use only a single transaction view; \proposed+CollapsedView1 combines edges from both views into a single directed graph while for \proposed+CollapsedView2 each transaction is reduced to a single undirected edge with both view weights as edge attributes; \proposed$-$EdgeWeights removes edge weights entirely. \textbf{Fusion variants} modify the integration mechanism: \proposed$-$Gating replaces gated fusion with concatenation, \proposed$-$ConsReg removes the consistency regularizer, and \proposed+CrossAttn replaces gating with bidirectional cross-attention between views.

The full \proposed~achieves the highest performance over all variants. Both single-view models deteriorate similarly, indicating that both perspectives are equally important. Both collapsed views result in a larger drop, highlighting the necessity of separately modeling buyer and seller views. Removing edge weights degrades performance below even single-view models, confirming that transaction scale plays a critical role in default prediction.

\looseness=-1
Among the fusion variants, replacing gating with concatenation or removing the regularizer both degrade performance while still outperforming single-view models. Replacing gating with cross-attention causes an even larger drop, confirming that preserving view-specific representations and adjusting only their relative importance is more effective than allowing views to modify each other.

\looseness=-1
Further, \proposed~is robust across hyperparameter choices: varying embedding dimension $d \in \{32, 64,$ $ 128, 256\}$, regularization weight $\lambda \in \{0.01, 0.05, 0.1, 0.2\}$, and edge scaling factor $\alpha \in \{1, 10, 50, 100, 200\}$, the NoHist AR ranges from 0.452 to 0.472 on the 2021 dataset, consistently outperforming the best baselines.

\begin{figure}[t]
\centering
\subfloat{
\includegraphics[width=0.95\linewidth]{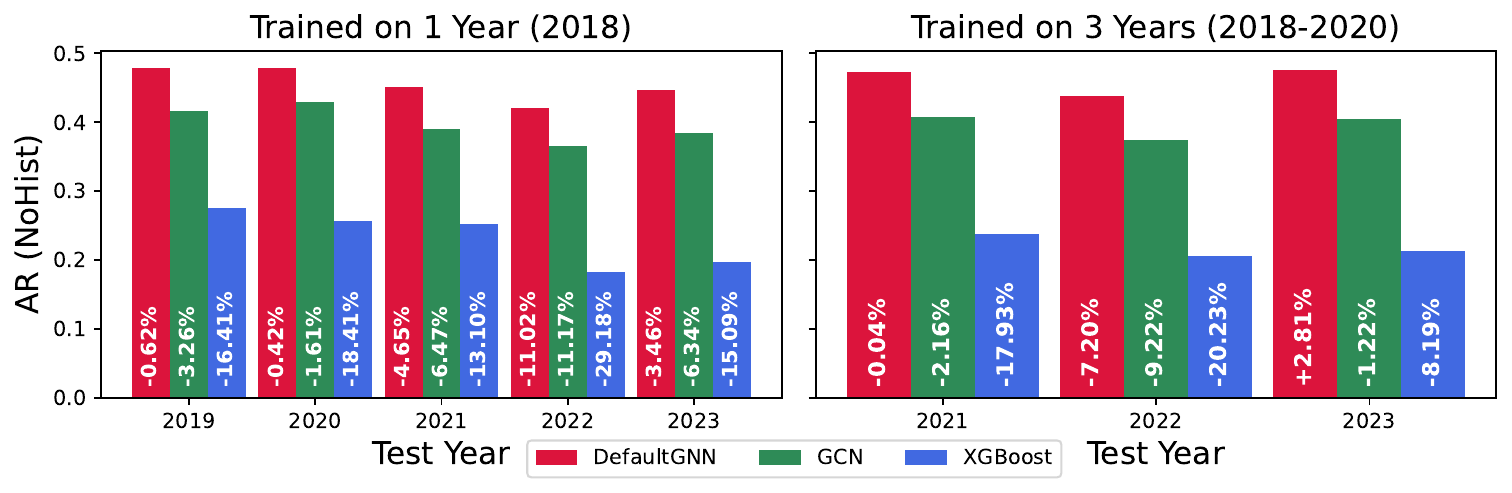}
}
\vspace{-3ex}
\caption{Comparison of model performances under OOT settings. Bar heights represent each model's AR score on the NoHist subset, while values in the bars show how much the OOT performance of each model diverges from the original performance, i.e., training and testing on the same year.}
\vspace{-3ex}
\label{fig:oot}
\end{figure}

\vspace{-2ex}
\subsection{Out-of-time Analysis (RQ3)} \label{oot}
To evaluate generalization in a realistic credit risk assessment setting where models are trained on historical data and tested on future years~\cite{statmodel, oot1, oot2, ar1}, we compare \proposed~against GCN and XGBoost in two out-of-time (OOT) configurations: (1) training on a single year (2018) and testing on 2019--2023, and (2) training sequentially on three years (2018--2020) and testing on 2021--2023.

As shown in Fig.~\ref{fig:oot}, \proposed~consistently outperforms baselines across both configurations. In the single-year setting, \proposed~maintains strong performance even as the training-test gap increases to five years, whereas GCN and XGBoost exhibit substantially larger degradation. In the three-year setting, sequential retraining allows \proposed~to closely match or even improve upon its in-year performance, demonstrating effective integration of new data over time. These results confirm that \proposed~generalizes well to future unseen data, a critical 
requirement for real-world credit risk models that must adapt to evolving economic conditions.

\subsection{Case Studies (RQ4)} \label{case}
\begin{figure}[t]
\centering
\includegraphics[width=0.85\linewidth]{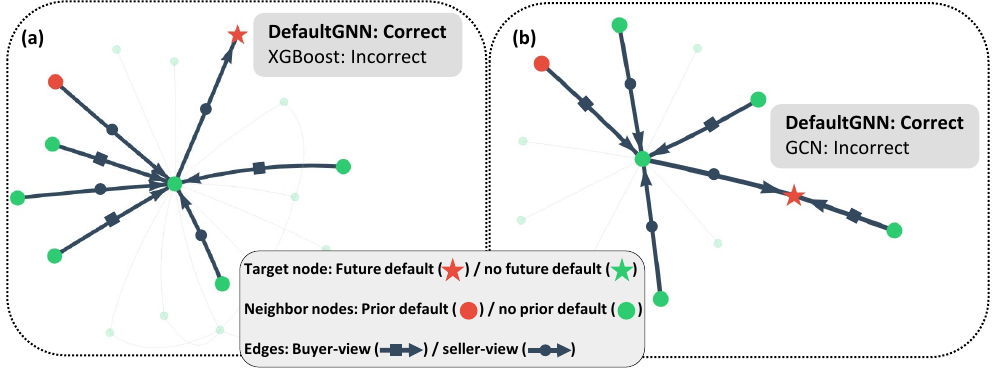}
\vspace{-3ex}
\caption{Case studies, where edges (transactions) and corresponding neighbors (trading partners) influential to the prediction of the target node (star-shaped) are highlighted.}
\vspace{-4ex}
\label{fig:casestudy}
\end{figure}

\looseness=-1
We present two case studies in Fig.~\ref{fig:casestudy} to illustrate how \proposed~identifies meaningful transaction-based risk signals beyond raw prediction performance. Specifically, for each target firm, we extract a two-hop local transaction subgraph from both buyer and seller views and attribute the prediction to individual transaction edges using integrated gradients~\cite{ig1, ig2, ig3}, which identifies the trading relationships and partners that most strongly drive the model's decision.

\looseness=-1
In Fig.~\ref{fig:casestudy}(a), \proposed~highlights a transaction with a previously defaulted firm located two hops away from the target firm, demonstrating its ability to capture indirect risk propagation. In contrast, XGBoost fails to incorporate this dependency and incorrectly predicts the firm as safe. In Fig.~\ref{fig:casestudy}(b), an influential relationship corresponds to a buyer-view transaction with a defaulted trading partner, which is then propagated to the target firm through a seller-view transaction. While ~\proposed~ correctly captures this signal through its dual-view design, a single-view GCN misses the risk and produces an incorrect prediction, underscoring the importance of modeling asymmetric transaction roles.

Beyond improving predictive accuracy, these visualizations offer clear and actionable insights by identifying influential trading partners and transaction links. Such interpretability can support financial institutions in understanding risk exposure, monitoring vulnerable relationships, and making informed credit decisions.

\vspace{-1ex}
\section{Practical Deployment}
\looseness=-1
To evaluate practical applicability, \proposed's predictions were validated in collaboration with Techfin Ratings, a licensed credit rating agency that operates a logistic regression-based credit scoring model (MIS) using real-time financial and operational data. \proposed's predicted default scores and MIS scores were used as input features to train a separate logistic regression model, assessing whether transaction network information provides complementary value to existing credit models. The evaluation covered 1,166,927 corporate firms and 197,397 sole proprietors using unseen data from 2018 onward, with the sequentially trained \proposed~model from Section~\ref{oot} (trained on 2018--2020) applied without retraining. In practice, default prediction scores are used by financial institutions to inform credit approval decisions, where firms below a given risk threshold are deemed eligible for lending. To simulate this scenario, a risk threshold was applied to each model's predicted scores to determine firm eligibility for credit approval. Integrating \proposed~increased the approval rate by 6.95 percentage points for corporate firms (51.57\% to 58.52\%) and 11.11 percentage points for sole proprietors (36.13\% to 47.24\%), while maintaining or reducing the default rate among approved firms (unchanged at 0.41\% for corporate firms; 0.14\% to 0.11\% for sole proprietors). These results suggest that transaction network-based risk signals complement existing credit models by expanding the pool of approvable firms without increasing portfolio risk, particularly benefiting SMEs and sole proprietors for whom traditional financial indicators are limited.

\vspace{-2ex}
\section{Conclusion}
\looseness=-1
In this work, we propose \proposed, a GNN-based framework for corporate default prediction that leverages buyer–seller transaction networks under realistic data constraints where financial statements are limited or unavailable. Through extensive empirical analysis of large-scale transactional data, we show that inter-firm transactions encode critical risk signals related to transaction role and scale. Guided by these findings, ~\proposed~ models transactions from dual buyer and seller perspectives while incorporating transaction magnitude, consistently outperforming existing baselines — particularly for firms without prior default history. Beyond predictive performance, the model provides interpretable insights that help identify influential trading partners, offering practical value for real-world credit risk assessment and monitoring. Further, \proposed's predictions complement an existing credit scoring model, greatly improving approval rates without increasing portfolio risk.

\smallskip
\noindent{\textbf{Acknowledgements. }}
This work was supported by the National Research Foundation of Korea (NRF) grants funded by the Korea government (MSIT) (RS-2024-00335098) and the Ministry of Science and ICT (RS-2022-NR068758), and by Douzone/Techfin Ratings.

\clearpage
\section*{GenAI Disclosure}
We acknowledge the use of LLMs (e.g., GPT-5, Claude) for limited assistance with (1) editing this paper for grammar, clarity, expression variation, and length reduction to meet page limits, and (2) minor refactoring/debugging of code used for plotting and visualization. All AI-assisted edits and code changes were reviewed and validated by the authors, and all core ideas, methods, experiments, and interpretations are original contributions of the authors.

\bibliographystyle{ACM-Reference-Format}
\balance
\bibliography{reference}

\end{document}